\documentclass{article}
\usepackage[preprint]{colm2025_conference}
\usepackage{multicol}
\usepackage{booktabs}
\usepackage{amssymb}
\usepackage{wrapfig}
\usepackage{graphicx}
\usepackage{microtype}
\usepackage{makecell}
\usepackage{subcaption} 
\usepackage{longtable}
\usepackage[T1]{fontenc}    
\usepackage{tablefootnote}
\usepackage{threeparttable}
\usepackage{xcolor}
\usepackage{pifont}
\usepackage{xcolor}
\newcommand{\cmark}{\ding{51}}
\newcommand{\xmark}{\ding{55}}
\newcommand{\pmark}{\ding{108}}
\definecolor{green}{RGB}{0,150,10}
\definecolor{blue}{RGB}{0,148,181}
\definecolor{orange}{RGB}{194,153,107}
\definecolor{HardBlue}{RGB}{0,45,120}

\usepackage[
    colorlinks=true,
    linkcolor=red,
    anchorcolor=green,
    citecolor=blue,
    urlcolor=HardBlue
]{hyperref}
\definecolor{HardBlue}{RGB}{0,45,120}
\hypersetup{
  colorlinks=true,
  urlcolor=HardBlue,
}
\usepackage{url}            
\usepackage{lmodern}
\usepackage{enumitem}

\usepackage{lineno}
\usepackage{tabularx}

\usepackage{booktabs}       
\usepackage{amsfonts}       
\usepackage{nicefrac}       

\definecolor{background-grey}{RGB}{220,220,220}
\definecolor{cell-green}{RGB}{221, 255, 225}  
\definecolor{cell-red}{RGB}{255, 224, 224}  
\definecolor{light-green}{HTML}{A2D9A2}
\definecolor{llight-green}{HTML}{C7EFCF}
\definecolor{light-red}{HTML}{FFD1D1}
\definecolor{light-orange}{HTML}{FFC9A3}

\usepackage{float}
\usepackage{amsmath}
\usepackage{marvosym}
\usepackage{graphicx}
\usepackage{colortbl}
\usepackage{multirow}
\usepackage{subcaption}
\usepackage{changes}
\usepackage{bookmark}
\usepackage{listings}
\lstdefinelanguage{Dialogue}{
  morekeywords={Influencer,Voter,rating},
  sensitive=false,
  morecomment=[l]{//},
}
\usepackage{natbib}
\usepackage{bibentry}
\nobibliography*
\usepackage[T1]{fontenc}

\usepackage{mdframed}
\usepackage{caption}
\usepackage{fancyhdr}
\usepackage{datetime}
\usepackage{adjustbox}
\usepackage{amssymb}
\usepackage{tcolorbox}
\usepackage{makecell}
\usepackage{diagbox}
\usepackage{multirow}      
\usepackage{array}         
\usepackage{fancyvrb}
\usepackage{fvextra}
\usepackage{pifont}
\usepackage{tikz}
\usepackage{wasysym}
\usepackage{xspace}
\usepackage{setspace}

\usepackage{soul} 

\definecolor{evidbgcolor}{HTML}{FFE6E6}
\definecolor{stepbgcolor}{HTML}{F0F0F0}
\definecolor{evidfgcolor}{HTML}{CC0000}

\usepackage{ulem}     

\definecolor{lightred}{RGB}{255,200,200}

\newcommand{\code}[1]{\texttt{#1}}
\newcommand{\framework}[1]{\textit{#1}}

\tcbuselibrary{listings,breakable}

\usepackage{geometry}
\newcommand{\AtwoE}{A\textsuperscript{2}E}
\newcommand{\AEtwo}{A\textsuperscript{2}E}
\newcommand{\ATP}{ATP}

\definecolor{Blue4Head}{RGB}{58,104,153}

\title{%
\begin{center}
  \begin{minipage}[c]{0.22\textwidth}
    \centering
    \includegraphics[width=3cm]{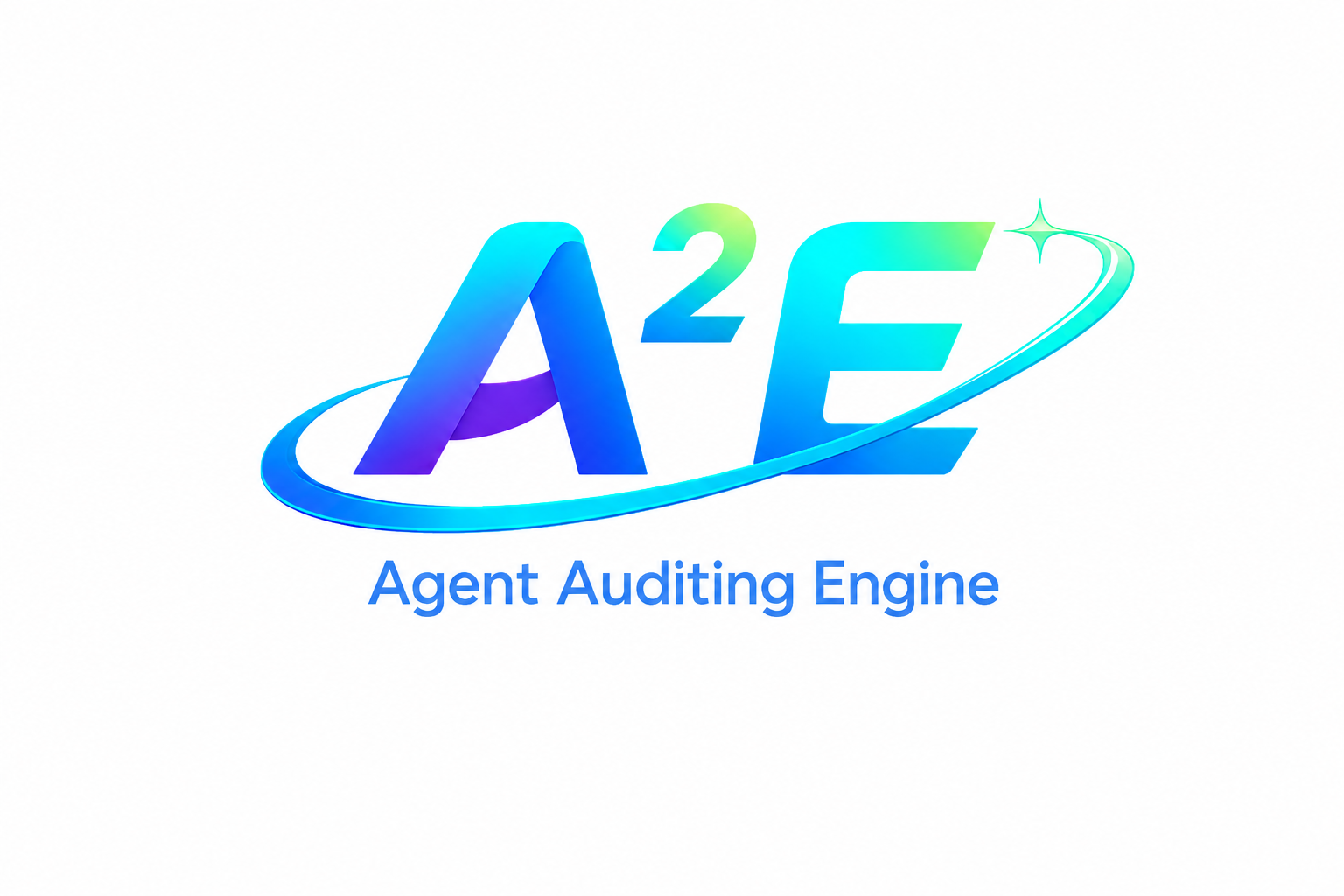}
  \end{minipage}
  \begin{minipage}[c]{0.7\textwidth}
    An End-to-End Agent Auditing Engine
  \end{minipage}
\end{center}
}

\author{%
  \makebox[\textwidth][c]{%
    Haoning Wang\textsuperscript{*}\quad
    Mingxun Zhang\textsuperscript{*}\quad
    Chenyue Yu\textsuperscript{*}\quad
    Yingjun Shang\textsuperscript{*}\quad
    Xing Xie\quad
    Xia Hu\quad
    Guanchu Wang\textsuperscript{\dag}\quad
    Na Zou\textsuperscript{\dag}%
  }\\[0.35em]
  \makebox[\textwidth][c]{Shanghai Artificial Intelligence Laboratory}\\[0.35em]
  \makebox[\textwidth][c]{%
    \normalfont\footnotesize
    \textsuperscript{*}Equal contribution.\quad
    \textsuperscript{\dag}Corresponding authors:
    \texttt{wangguanchu@pjlab.org.cn}, \texttt{zouna@pjlab.org.cn}.%
  }\\[0.4em]
  \makebox[\textwidth][c]{\includegraphics[height=1.5em]{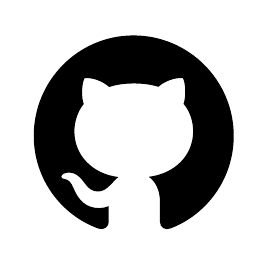}\;\href{https://github.com/datamllab/AE2.git}{\textsf{\bfseries{\textcolor{HardBlue}{https://github.com/datamllab/A2E}}}}}%
}

\begin{document}

\hypersetup{
    linkcolor=black,
    filecolor=black,      
    urlcolor=blue,
    citecolor=blue,
}

\maketitle

\begin{abstract}
  With the rapid advancement of large language models (LLMs), harnesses have become essential infrastructure for deploying agents across a wide range of domains. The fast-evolving harness ecosystem has also made rigorous capability evaluation increasingly important. However, efficiently building an end-to-end, systematic, and comprehensive evaluation pipeline remains a significant challenge. To address this challenge, we introduce \AtwoE{} (Agent Auditing Engine), an end-to-end evaluation engine designed for agent harnesses.
 \AtwoE{} leverages our newly proposed Agent Task Protocol (\ATP{}) to enable the rapid integration of evaluation tasks with different harnesses. Through an automatically instrumented Monitor, it captures and generates standardized execution traces during experiments. In the Evaluation stage, \AtwoE{} systematically assesses harness capabilities using a suite of multidimensional metrics. Compared with correctness alone, these metrics provide a more fine-grained characterization of differences among harnesses in execution efficiency, tool use, task planning, and error recovery. Experiments conducted with \AtwoE{} further reveal that model–harness combinations exhibit substantial performance variation across different types of tasks, and that no single combination consistently outperforms all others across every task. These findings not only demonstrate the necessity of systematic evaluation but also provide useful guidance for the co-evolving of models and harnesses.
  \end{abstract}

\begingroup
\captionsetup{font=footnotesize}
\vspace{-5mm}
\vfill
\noindent\begin{minipage}{\textwidth}
  \centering
  \begin{minipage}[c]{0.577\textwidth}%
    \centering
    \includegraphics[width=\linewidth]{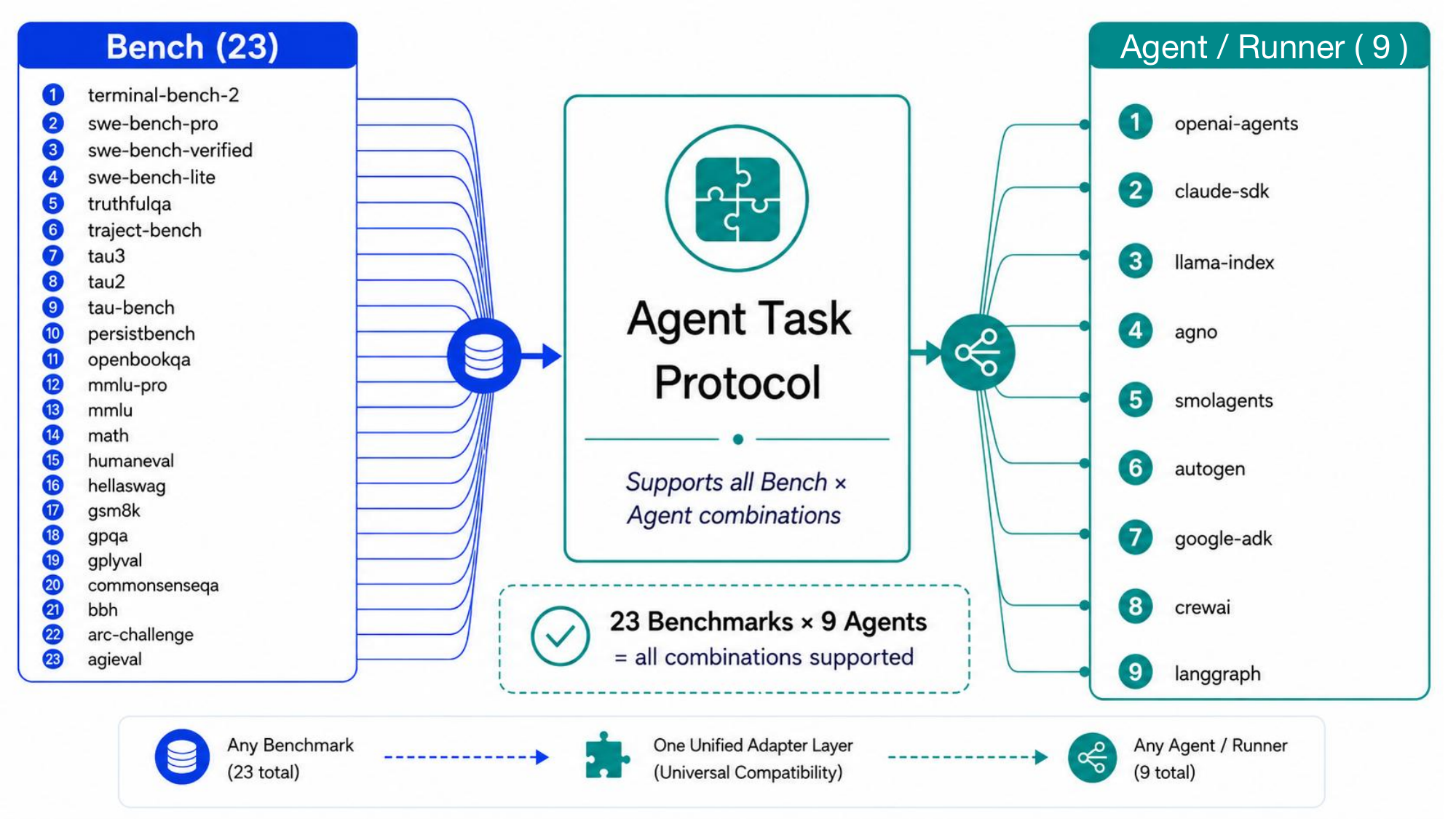}
  \end{minipage}\hspace{0.035\textwidth}%
  \begin{minipage}[c]{0.314\textwidth}%
    \centering
    \includegraphics[width=\linewidth]{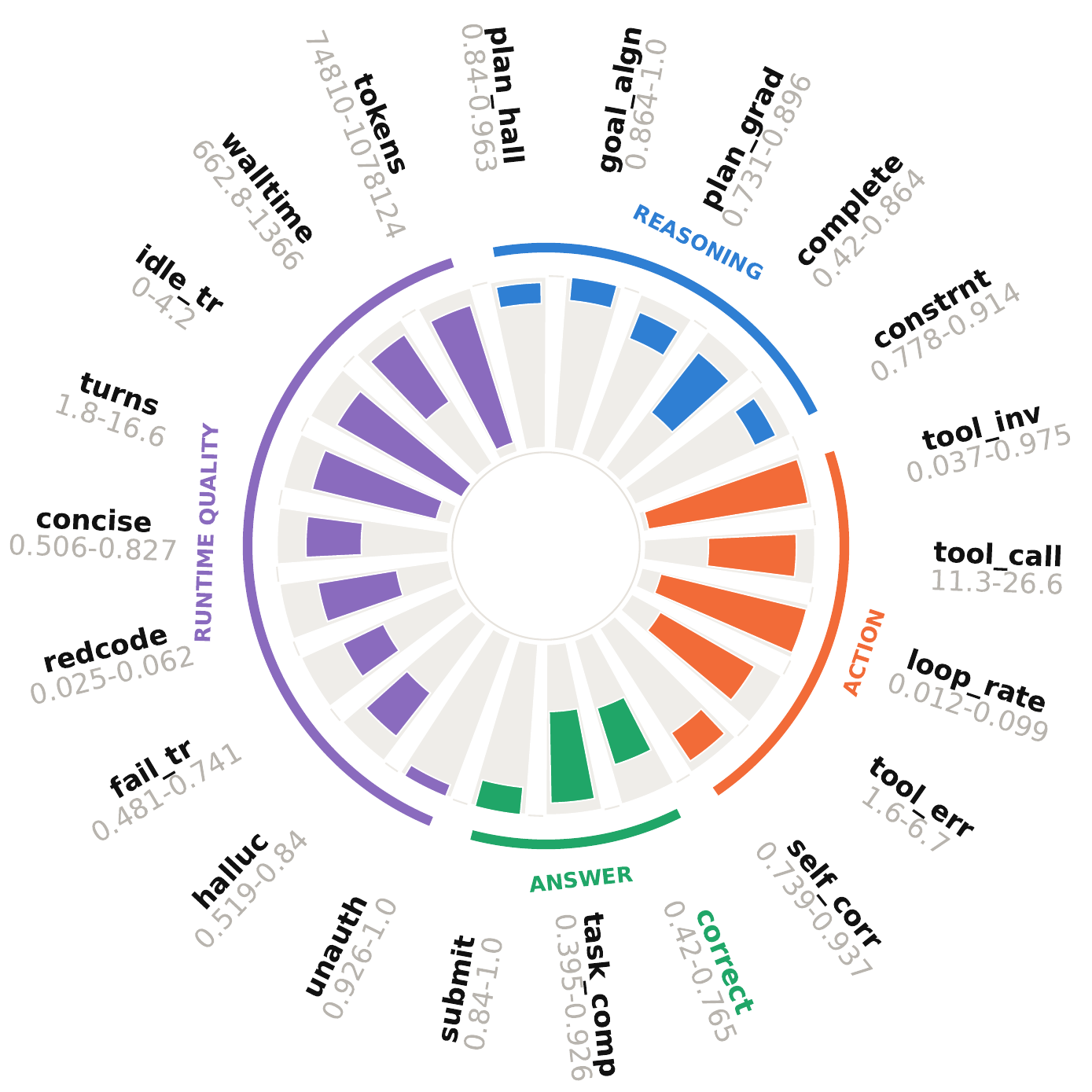}
  \end{minipage}

  \vspace{0.4em}
  \begin{minipage}[t]{0.577\textwidth}%
    \centering\footnotesize (a) Unified adapter for all $m \times n$ pairings
  \end{minipage}\hspace{0.035\textwidth}%
  \begin{minipage}[t]{0.314\textwidth}%
    \centering\footnotesize (b) Harness capability boundaries
  \end{minipage}

  \captionof{figure}{\AtwoE{} provides end-to-end evaluation over an $m \times n$ benchmark--harness grid. \textbf{(a)}~An Agent Task Protocol lets each of the 23 benchmarks be paired with each of the 9 agent frameworks, so no per-combination integration code is written. \textbf{(b)}Each petal shows, for one metric, the span from the worst to the best of the nine harnesses after averaging over the 23 benchmarks. Petals are grouped by Reasoning, Action, Answer, and Runtime Quality. \texttt{correctness} spans only about $0.42$--$0.77$, indicating limited differences in final-answer accuracy, while planning, tool use, and efficiency vary much more widely across harnesses.}
  \label{fig:intro-capability}
\end{minipage}
\endgroup

\newpage

{\small \tableofcontents} 

\newpage
\hypersetup{
    linkcolor=red,
    filecolor=black,      
    urlcolor=blue,
    citecolor=blue,
}

\setcounter{section}{0}
\section{Introduction}
\label{sec:introduction}
 
As large language models improve, overall system performance increasingly depends on the agent harness. The harness determines the system prompts, tool interfaces, context management, and execution policies.
Evaluating the underlying model alone does not fully capture the capabilities of a deployed agent system. Reproducible harness-level evaluation and fine-grained trajectory collection are therefore increasingly important.
However, existing frameworks\citep{aisi2024inspect,arize2024phoenix,xi2025agentgym} typically address only part of this workflow.
Inspect AI~\citep{aisi2024inspect} provides comprehensive support for benchmark orchestration, sandboxed execution, and scoring, but integrates external harnesses primarily through harness-specific adapters and model API proxies.
Such integration requires continuous maintenance as harness interfaces and event formats evolve, and may alter generation parameters, invocation paths, or other aspects of native execution, potentially reducing trajectory fidelity.
In contrast, OpenInference-based systems such as Phoenix~\citep{arize2024phoenix} provide standardized observability for native agent executions but do not constitute an end-to-end benchmark runner with task orchestration and result verification, leaving users to integrate benchmarks, harnesses, and tracing infrastructure themselves.
Although both ecosystems are extensible, extending them to new benchmarks, harnesses, or trajectory semantics often requires substantial framework-specific engineering.
This motivates a lightweight, minimally invasive, and harness-agnostic substrate that unifies benchmark execution with faithful trajectory collection.
\begin{table}[H]
\centering
\caption{\cmark: fully supported; \pmark: partially supported; \xmark: not supported.}
\label{tab:framework_comparison}
\vspace{-5pt}
\footnotesize
\setlength{\tabcolsep}{3pt}
\renewcommand{\arraystretch}{1.0}
\begin{tabular}{lcccccc}
\toprule
System & \makecell{Task\\Orchestration} & \makecell{Native\\Execution} & \makecell{Trajectory\\Collection} & \makecell{Lifecycle\\Evaluation} & \makecell{Persistent\\Storage} & \makecell{Harness\\Agnostic} \\
\midrule
Inspect AI & \cmark & \xmark & \cmark & \pmark & \xmark & \pmark \\
Phoenix & \xmark & \cmark & \cmark & \xmark & \pmark & \cmark \\
\AtwoE{} & \cmark & \cmark & \cmark & \cmark & \cmark & \cmark \\
\bottomrule
\end{tabular}
\vspace{-8pt}
\end{table}
To address these limitations, we present \AtwoE{}, a lightweight end-to-end engine that unifies task composition, trajectory monitoring, and lifecycle-aligned evaluation within a single evaluation stack.
\AtwoE{} consists of three layers. First, its Task Layer, built on the Agent Task Protocol (\ATP{}), decouples benchmarks from agent harnesses and allows them to be composed independently.
This modular abstraction enables new benchmarks and harnesses to be integrated with minimal adapter code, avoiding pairwise implementations for every benchmark–harness combination.
Second, the Monitor Layer builds on \textit{OpenInference} to capture agent execution traces using OpenTelemetry-compatible spans~\citep{opentelemetry}.
The resulting standardized trajectories preserve structured events across model invocations, tool calls, and execution stages, while remaining interoperable with the broader \textit{OpenTelemetry} observability ecosystem.
Third, the Evaluation Layer introduces Lifecycle-Aligned Evaluation, in which each metric is registered under a specific execution stage and a fine-grained evaluation dimension. This design makes the evaluation framework extensible while ensuring that metrics are aligned with the stage of the agent lifecycle they are intended to assess. \AtwoE{} further stores trajectories, metric definitions, and evaluation results in a database rather than relying on standalone log files, enabling incremental, persistent, and longitudinal evaluation across runs. By retaining only the essential abstractions required by these three layers, \AtwoE{} achieves a substantially smaller core implementation than general-purpose systems such as Inspect AI~\citep{aisi2024inspect} and Phoenix~\citep{arize2024phoenix}, while providing an integrated and extensible foundation for harness-level evaluation.
 
Using \AtwoE{}, we conduct a large-scale study of harness–model interactions and obtain two key findings. First, there is no universally dominant harness–model configuration. As shown in Figure~\ref{fig:cross_benchmark_harness_comparison}, the configurations occupying the success–efficiency frontier vary substantially across GDPVal\citep{patwardhan2025gdpval}, MMLU-Pro\citep{wang2024mmlupro}, and $\tau^3$-bench\citep{yao2024taubench,barres2025tau2bench}: a configuration that achieves high task success or favorable token efficiency on one benchmark may perform poorly on another. The best-performing harness also changes across models and tasks, demonstrating that harness effectiveness is inherently model- and task-dependent rather than globally rankable. Second, endpoint correctness alone has limited resolution for distinguishing harnesses. Under our Lifecycle-Aligned Evaluation, we score DeepSeek-V4-Pro with nine harnesses across the matched campaign and compare metrics that span planning, tool use, final answers, and operational quality by the range of harness. As shown in Figure~\ref{fig:intro-capability}, these signals diagnose the full trajectory from deliberation through action to the answer, including cost and safety, rather than only whether the final output is correct. Several process and operational metrics open wider petals than \texttt{correctness}, whose harness means remain narrowly concentrated: frameworks that look similar on the outcome layer still diverge in how they plan, invoke tools, and spend compute.
 
\section{Overview}
\label{sec:overview}

\begin{figure*}[t]
    \centering
    \includegraphics[width=\textwidth]{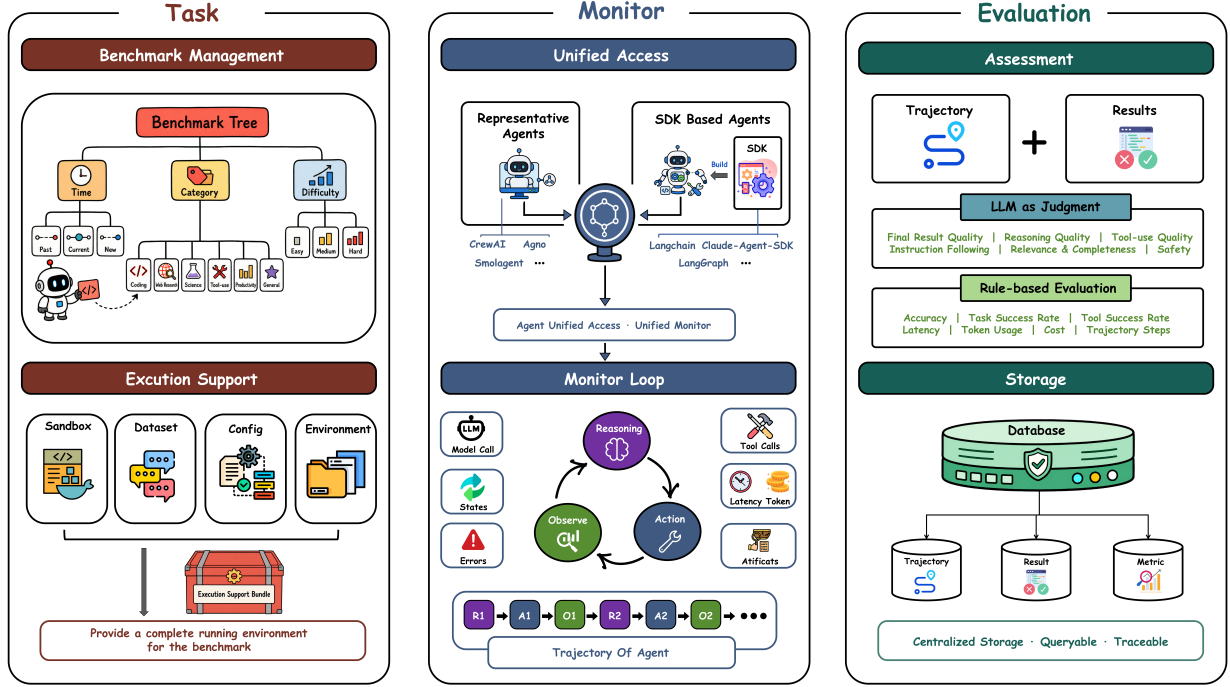}
    \caption{System overview. \emph{Task} integrates benchmark management
    and execution support, \emph{Monitor} provides unified agent access
    and instruments the runtime loop, and \emph{Evaluation} performs
    multi-dimensional assessment with centralized result storage.}
    \label{fig:overview}
\end{figure*}

Our engine provides an end-to-end infrastructure for organizing,
executing, monitoring, and evaluating LLM agents across heterogeneous
benchmarks. As illustrated in Fig.~\ref{fig:overview}, the system
consists of three major components: Task Layer, Monitor Layer, Evaluation Layer. Together, these components form a closed
workflow that connects benchmark preparation with agent execution,
trajectory collection, multi-dimensional evaluation, and result
analysis.

\subsection{Framework}
\label{sec:overview-components}

\noindent\textbf{Task Layer.}
The \emph{benchmark management} block provides a unified mechanism for
organizing diverse agent benchmarks. Benchmarks are maintained in a
hierarchical benchmark tree and indexed along three dimensions: time,
category, and difficulty. The time dimension separates past, current,
and new benchmarks, so that evaluation can distinguish saturated
benchmarks from recently released ones. The category dimension covers
representative domains such as coding, web research, science, tool use,
productivity, and general-purpose tasks. The difficulty dimension labels
tasks as easy, medium, or hard, supporting fine-grained comparison
beyond a single aggregate score.

The \emph{execution support} block packages the resources required to
reproduce each benchmark into an execution-support bundle. A bundle
contains the sandbox definition, task dataset, experiment configuration,
and runtime environment, and together these four elements provide a
complete running environment for the benchmark. To standardize how tasks
are presented to agents and how runs are recorded, benchmarks in this
bundle follow \ATP{}. During execution, the
selected benchmark and agent are instantiated together in the sandbox,
ensuring that different agents are evaluated under consistent and
reproducible conditions. This abstraction hides benchmark-specific setup
details and allows new benchmarks to be integrated without modifying the
remaining experiment and evaluation pipeline.

\noindent\textbf{Monitor Layer.}
The \emph{unified access} block exposes a single entry point for two
classes of agents. Representative agents, such as CrewAI, Agno, and
Smolagent, are ready-to-run systems that are invoked directly, whereas
SDK-based agents are built with popular development kits and frameworks,
including LangChain, Claude-SDK, and LangGraph. This layer
normalizes their heterogeneous invocation interfaces into a common
agent-access abstraction and attaches the same unified monitor to every
agent. Consequently, the same execution and observation pipeline applies
regardless of how an agent is implemented.

The \emph{monitor loop} block instruments the iterative
reasoning--action--observation cycle that the agent follows during an
experiment. At every step the monitor captures model calls, state
transitions, errors, tool calls, latency and token consumption, and
generated artifacts. Each run is recorded as an ordered event sequence
$R_1 \rightarrow A_1 \rightarrow O_1 \rightarrow R_2 \rightarrow \cdots$,
which preserves both the final task outcome and the intermediate
decisions that produced it. The collected run information and traces are
streamed to the centralized server, which provides a common data
interface for subsequent evaluation, storage, and visualization. This
automatic instrumentation minimizes the integration effort required from
benchmark and agent developers while enabling fine-grained analysis of
agent behavior.

\noindent\textbf{Evaluation Layer.}
The \emph{assessment} block takes both the execution trajectory and the
final results as input. Outcome evaluation determines whether the agent
completes the task correctly, whereas trajectory evaluation examines how
the agent arrives at its answer; combining the two avoids relying
exclusively on final-answer correctness and exposes inefficient, unsafe,
or otherwise undesirable behaviors that remain hidden in aggregate
success rates. Assessment is carried out by two complementary families
of evaluators. Rule-based evaluation computes accuracy, task success
rate, tool success rate, latency, token usage, cost, and the number of
trajectory steps. LLM-as-judge evaluation scores qualitative dimensions
that rules cannot express, namely final-result quality, reasoning
quality, tool-use quality, instruction following, relevance and
completeness, and safety.

The \emph{storage} block persists the outputs of this process.
Evaluators retrieve the required traces and outcomes from the server and
write their scores and annotations back as structured records, which the
centralized database organizes into trajectory, result, and metric
stores. This centralized design keeps experiment results queryable and
traceable across benchmarks, agents, models, and configurations. On top
of this layer, the user interface presents individual trajectories and
scores as well as aggregated comparisons, allowing users to inspect
failures, analyze agent behavior, and compare experimental results
through a consistent view.

\subsection{Pipeline}
\label{sec:overview-pipeline}

\begin{figure*}[t]
    \centering
    \includegraphics[width=0.8\textwidth]{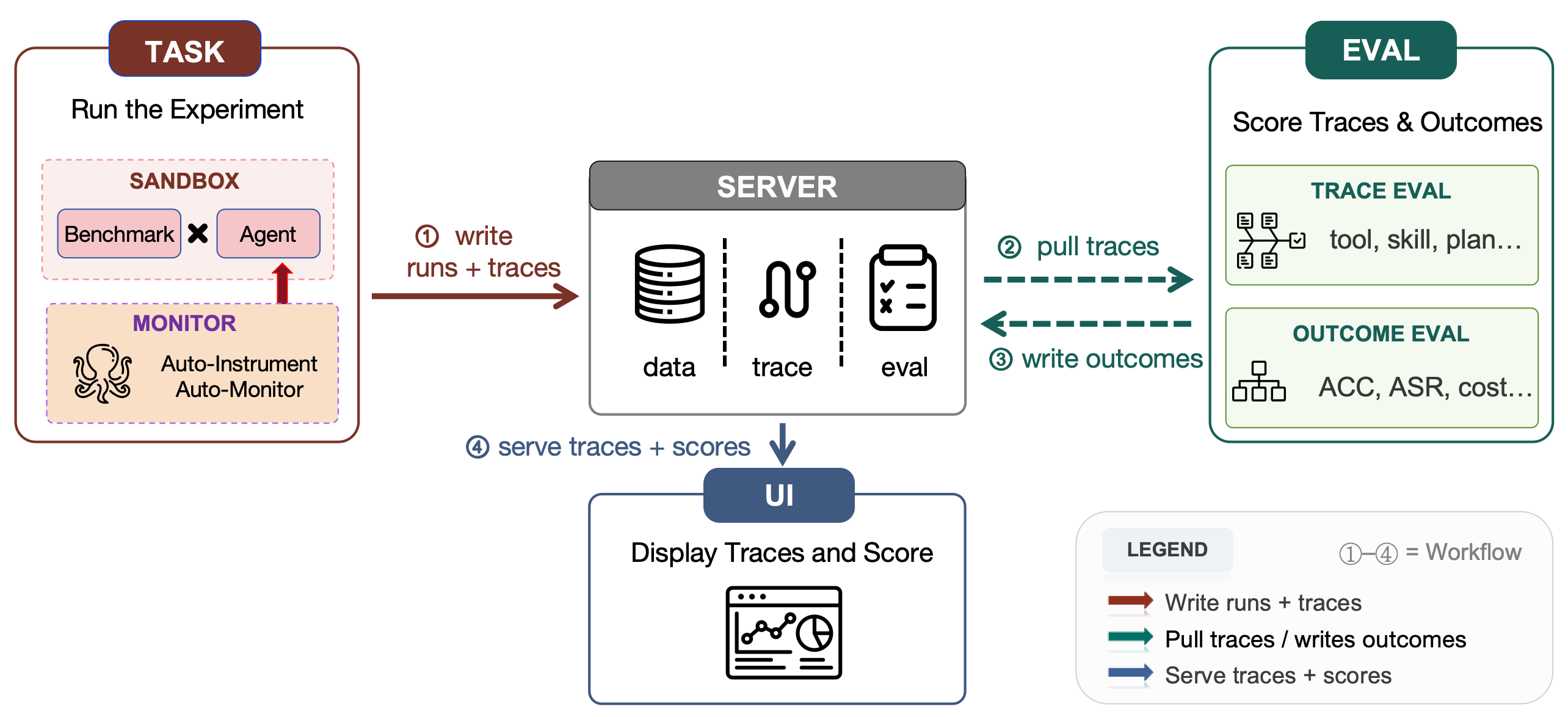}
    \caption{Runtime workflow and data flow. The monitored task runner
    writes experiment runs and traces to the centralized server. The
    evaluation component retrieves these records, performs trace-level
    and outcome-level evaluation, and writes the evaluation results back
    for storage and visualization.}
    \label{fig:runtime-workflow}
\end{figure*}

Fig.~\ref{fig:runtime-workflow} presents the runtime pipeline, which is organized into four loosely coupled components: Task, Server, Evaluation, and UI. During execution, a benchmark and an agent run inside an isolated sandbox. The monitoring module automatically instruments their interaction, collecting run metadata, tool calls, intermediate states, and execution traces without requiring custom logging code for each benchmark or agent. The resulting records are continuously written to the centralized server. The evaluation component then retrieves the traces and task outputs, computes trace-level and outcome-level metrics, and writes the evaluation results back to the server.

This separation keeps the system logic clear and allows each component to be developed, tested, and extended independently. In particular, the server acts as the central coordination point for all experiment data, including configurations, instrumented traces, task outputs, and evaluation records. As a result, the evaluator and the UI do not need direct access to the task runtime.

In our implementation, the UI is intentionally read-only: it queries the server database to display execution traces, scores, and aggregated experiment results, but does not launch experiments or modify evaluation records. This design avoids coupling visualization with execution and evaluation logic, reduces the risk of unintended state changes, and ensures that the displayed results remain consistent with the records stored on the server.

\section{Monitor Layer}
\label{sec:monitor-layer}

The Monitor layer reveals how an agent completes a task by recording intermediate activities such as reasoning chains, model interactions, tool use, and skill invocation. It consists of instrumentation that captures these activities and span-based traces that organize the resulting observations. Together, they support the analysis of both execution outcomes and the processes that produce them.

\subsection{Instrumentation}
\label{sec:monitor-instrumentation}

Monitor instruments behaviors that are usually hidden behind an agent's final response and turns them into observable trace data. Instead of requiring each agent to describe its own execution manually, the monitoring process observes the natural execution points exposed by the underlying framework. It consists of three connected parts that separate semantic interpretation, runtime recording, and framework integration.

\noindent\textbf{Semantic layer.}
The semantic layer defines the principal forms of agent behavior, including agent, chain, model call, tool, and skill. It gives these activities consistent meanings and identifies relevant information such as their inputs, outputs, and identities. This vocabulary establishes the conceptual boundary of each operation, allowing traces to be interpreted in terms of agent behavior rather than low-level runtime events.

\noindent\textbf{Span layer.}
The span layer records each recognized behavior as a bounded operation. It captures when an action begins and ends, the context in which it occurs, and whether it succeeds or fails. It also preserves the relationship between a behavior and the operation that initiated it. An agent run is consequently represented as a collection of meaningful and related activities rather than a flat sequence of logs.

\noindent\textbf{SDK layer.}
The SDK layer connects this observational model to different agent frameworks and model interfaces. It maps framework-specific mechanisms, such as model calls, tool execution, workflow transitions, and asynchronous operations, onto the semantic and span representations above. By containing these differences within dedicated adapters, the monitoring model remains stable as support for new frameworks is introduced.

These three parts operate as a continuous process. The SDK layer detects an activity in the agent runtime and translates its framework-specific representation; the semantic layer determines what the activity means; and the span layer records when it occurs and where it belongs in the execution hierarchy. Each recorded operation therefore answers three basic questions: what the agent did, when it happened, and which preceding operation led to it. Because framework-specific differences are resolved before the trace is organized, they do not obscure the higher-level execution flow. At the same time, temporal boundaries and parent--child relationships preserve the original structure of the run instead of reducing it to a flat event sequence. This division of responsibilities is what makes the resulting trace clear as both a chronological record and a causal account of agent behavior.

This organization also supports extension through a shared foundation with specialized adaptations. When a new framework is introduced, it can retain the existing semantic vocabulary and trace structure while adding only the interpretation required for its own execution model. Monitoring coverage can therefore grow without repeatedly redefining how agent behavior is represented, and traces remain understandable as the system evolves.

\subsection{Span-Based Traces}
\label{sec:span-based-traces}

Monitor adopts the OpenTelemetry span model to preserve the internal structure of an agent run. A span represents an operation with a start time, an end time, a status, and contextual information. Related spans share a trace identity and form a complete execution record, while parent--child relationships describe how one operation leads to another. The highest-level span represents the overall run, and its descendants capture progressively more specific activities.

This structure closely matches agent execution. A top-level agent span may contain reasoning chains, model calls, and tool or skill invocations. A model call can trigger a tool, while the tool result may return to the surrounding chain and influence the agent's next decision. These activities remain individually observable, but their relationships place them within the wider execution process. The trace therefore records both the sequence of actions and their causal organization.

Span-based tracing provides information that cannot be recovered from the final response alone. Span duration helps identify delays, while status and context help locate failures. Because operations are recorded separately, traces also make it possible to compare where different runs spend time or diverge in their behavior. More broadly, a trace shows which reasoning path was followed, when the model was consulted, and which external capabilities were used, enabling evaluation of both the result and the execution process.

\section{Task Layer}
\label{sec:task-layer}
The Task Layer introduces the Agent Task Protocol (\ATP{}), a shared interface that separates benchmark adaptation from agent harness execution.
\ATP{} provides a common representation for tasks across different benchmarks and enables agent harnesses to interact with them through a unified interface.
This section presents the design of \ATP{}, describes how benchmarks and agent harnesses are integrated through the protocol, and explains how the Task Layer supports their execution across different task settings.

\begin{figure*}[t]
    \centering
    \includegraphics[width=0.8\textwidth]{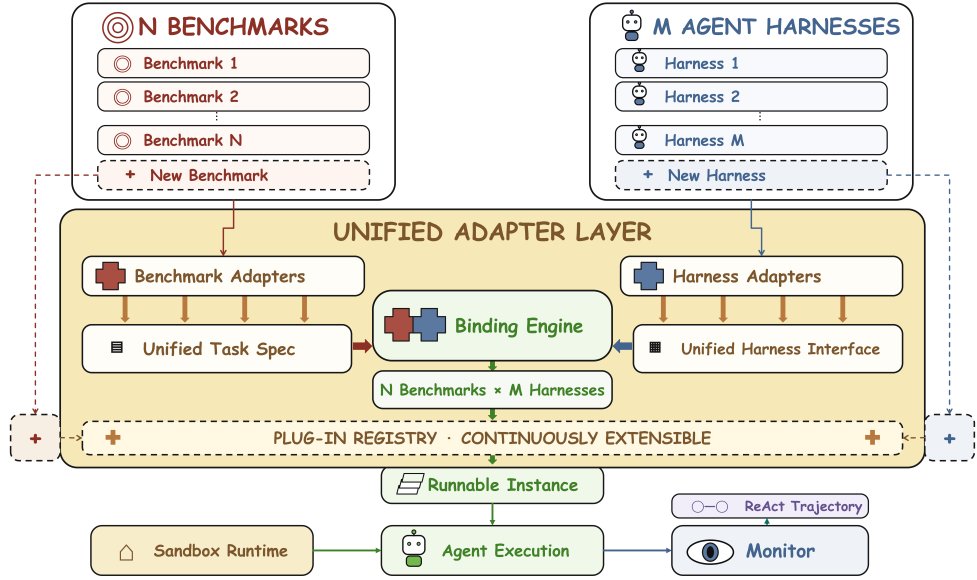}
    \caption{Overview of the Task Layer and the Agent Task Protocol (\ATP{}).
    Benchmark adapters and harness adapters map heterogeneous benchmarks and agent harnesses into unified interfaces.
    The binding engine composes them into $N \times M$ executable combinations through an extensible plug-in registry.
    Each bound instance is then executed with the corresponding runtime and monitored to produce an agent trajectory.}
    \label{fig:task}
\end{figure*}

\subsection{Agent Task Protocol (\ATP{})}
\label{sec:agent-task-protocol}

The task layer connects an agent harness to a benchmark and records one agent trajectory.
Agent harnesses expose different model clients, tool objects, and control loops.
Benchmarks define different instructions, states, tools, and execution environments.
A separate adapter for each harness-benchmark pair would mix benchmark logic with harness logic.
Recent agent systems also separate task interfaces from agent execution~\citep{gioacchini2024agentquest,bandel2026general,lacoste2026cube}.

Within\AtwoE{}, we call this shared interface the \emph{Agent Task Protocol} (\ATP{}).
\ATP{} defines the task representation, agent-facing interface, and execution record used by the task layer.
It separates benchmark adaptation from agent harness execution.
\ATP{} is an internal software protocol rather than a network protocol.

\AEtwo{} implements \ATP{} with four task-layer objects.
A benchmark adapter creates a \code{TaskInput} and an \code{AgentBinding}.
The task input stores the instruction, state, expected actions, expected outputs, metadata, and optional sandbox specification.
The binding provides tool schemas, tool execution, and prompt construction.
An \code{AgentRunner} is configured with the binding and runs each task input.
It returns a \code{TaskTrace} with the run status, final answer, and ordered tool calls.
These objects give benchmark adapters and agent harnesses a shared boundary.
Figure~\ref{fig:task} distinguishes observation from evaluation.

\subsection{Agent Harnesses \& Benchmarks}
\label{sec:agent-harness-benchmarks}
An agent harness is the task-layer integration of an agent framework.
It contains the framework-specific model client, control loop, tool conversion, and output normalization.
The current registry contains harnesses for \framework{Agno}, \framework{AutoGen AgentChat} \citep{wu2023autogen}, \framework{CrewAI} \citep{duan2024exploration}, \framework{Google ADK} \citep{huizenga2025agent}, \framework{LangGraph} , \framework{LlamaIndex} \citep{Liu_LlamaIndex_2022}, \framework{OpenAI Agents SDK}, \framework{Smolagents}, and the \framework{Anthropic Python SDK}.
\framework{AutoGen AgentChat} uses an isolated environment because its dependencies conflict with the main task environment.

Each registered agent harness follows \ATP{} through its \code{AgentRunner}.
The runner converts \ATP{} tool schemas into native framework objects.
It also configures the model client and keeps the framework control loop inside the harness.
The runner then converts the native result into a \code{TaskTrace}.
This design lets a benchmark adapter work with different registered agent harnesses.
Registry support does not imply that each framework--benchmark pair has passed end-to-end validation.

On the benchmark side, an adapter converts each source item into a \code{TaskInput}.
It also creates the \code{AgentBinding} required by the harness.
This report groups $23$ benchmarks into four task areas.
The groups define the scope of this report.
They do not replace the dataset kinds used by the registry.

The four task areas represent different forms of agent work.
Coding tasks cover code generation and repository modification.
Conversational tasks cover question answering, reasoning, and tool-based dialogue.
Research tasks focus on scientific questions that require specialized knowledge.
Computer-use tasks cover digital work and command-line interaction.

Across these four areas, dataset adapters support text, tool-use, and sandbox tasks.
Text tasks provide instructions and expected outputs.
Tool-use tasks add callable tools and an initial state.
Sandbox tasks add a container image, a working directory, and task setup.
Each dataset adapter converts its source items into \ATP{} objects.
Therefore, these execution differences do not change the boundary to the agent harness.

\subsection{Trajectory Generation}
\label{sec:trajectory-generation}

Trajectory generation starts with a benchmark key, an agent harness key, and a model name.
The registry resolves the dataset loader, binding, agent runner, and SDK mapping.
The CLI loads the candidate split and samples $40$ tasks without replacement by default.
The user can select another sample size or provide a random seed.
If no seed is provided, the CLI creates one.

The CLI next creates the \code{AgentBinding} and the selected \code{AgentRunner}.
For each sampled item, it builds a \code{TaskInput} and invokes the runner.
A sandbox task also receives a live environment through its task state.
The agent harness then runs its control loop and returns a \code{TaskTrace}.

The task trace is the normalized trajectory record.
It stores the run status, turn count, ordered tool calls, final answer, elapsed time, and raw framework output.
Runtime instrumentation records framework calls as a span tree.
A trace identifier links the task trace to this span tree when the execution path provides one.
This link preserves both the normalized result and the framework-level execution detail.

The task layer provides two trajectory generation paths.
The native \code{ExperimentRunner} creates one root span for each task and returns task traces directly.
The CLI path uploads the sampled tasks and passes a task function to \code{run\_experiment}.
Both paths follow \ATP{}.
They differ in orchestration and root-span ownership.

Each CLI run receives a unique run identifier.
The stored metadata records \code{agent\_framework}, \code{model}, \code{sdk}, the dataset key, the sample seed, and the selected task identifiers.
These fields identify the harness, benchmark, model, and sample used to generate the trajectories.
Reproduction also requires the same dataset version, model endpoint, and run settings.

\begin{figure*}[t]
    \centering
    \includegraphics[
        width=\textwidth,
        keepaspectratio
    ]{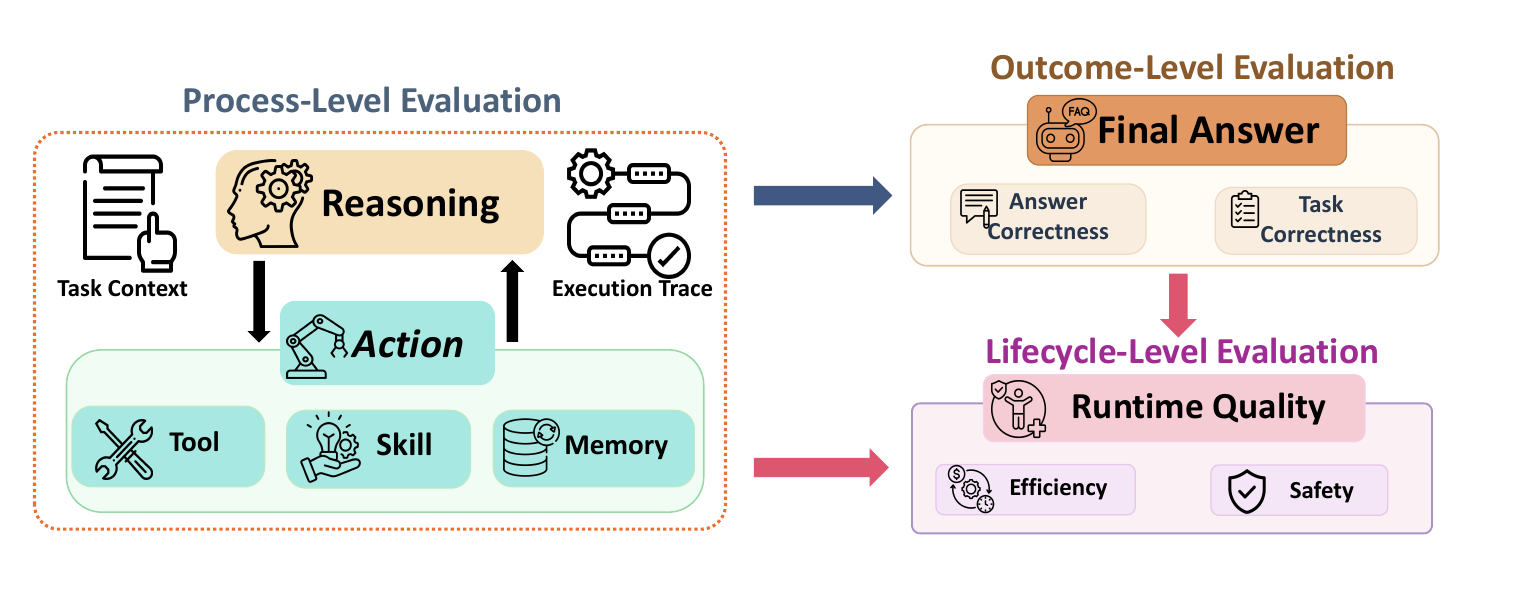}
    \caption{
        Overview of the execution-aligned agent evaluation framework.
        Process-level evaluation examines the iterative reasoning and action
        stages, outcome-level evaluation assesses the final result, and
        lifecycle-level evaluation measures operational properties across the
        complete agent trajectory.
    }
    \label{fig:agent_metric_taxonomy}
\end{figure*}

\section{Evaluation Layer}
\label{sec:evaluation}
Existing agent evaluations often reduce an interaction trajectory to task success or final-answer quality. While useful for high-level comparison, this outcome-only view has three limitations that our design addresses. First, it offers little diagnostic signal: when a run fails, one cannot tell whether the breakdown occurred in reasoning, tool use, memory, answer generation, or system operation. Second, metric catalogs are typically closed and hard to extend: adding a new property often requires modifying the benchmark runner or the harness. Third, trajectories are usually stored as isolated text or JSON files, making it difficult to aggregate, re-evaluate, or scale across many runs. Recent work has begun to address the first point through progress-based and trajectory-level analysis~\cite{ma2024agentboard}. Our framework goes further by embedding these insights into a unified architecture that is lifecycle-aligned, extensible, and database-backed. The metric catalog described below combines LLM-based evaluators for semantic and behavioral properties with deterministic metrics computed from execution traces, covering reasoning, tool use, memory, final-answer correctness, task completion, and operational properties such as token usage, cost, latency, safety, and prompt-injection resilience. This catalog is not intended as a complete definition of agent quality; rather, it demonstrates how the proposed architecture organizes, executes, stores, and aggregates heterogeneous metrics across the full agent lifecycle.

\subsection{Lifecycle-Aligned Evaluation}
\label{sec:lifecycle_metric_organization}

We organize evaluation metrics according to their positions in the agent
execution lifecycle, as illustrated in Figure~\ref {fig:agent_metric_taxonomy}. The resulting taxonomy contains four evaluation stages.

\textbf{Reasoning} \citep{sun2026agent} evaluates how the agent interprets the task and constructs
its solution process. It is further divided into \textbf{Task},
\textbf{Flow}, and \textbf{Logical} dimensions, corresponding to objective
understanding, planning completeness, and reasoning coherence.

\textbf{Action} \citep{maharana2024evaluating} evaluates how reasoning is converted into interactions with the execution environment. Its \textbf{Tool}, \textbf{Skill}, and
\textbf{Memory} dimensions examine tool-use behavior, capability application,
and the faithful use of previously available context, respectively.

\textbf{Final Answer} \citep{zhou2024webarena} evaluates the result produced after execution.
\textbf{Answer Correctness} measures the semantic quality of the response,
whereas \textbf{Task Completion} determines whether the underlying task
reaches a valid terminal state.

Finally, \textbf{Runtime Quality} captures properties that span the
complete trajectory rather than a single execution step. It contains
\textbf{Efficiency} and \textbf{Safety}, which measure resource consumption
and operational risks throughout the agent lifecycle.

This organization separates three complementary evaluation scopes:
process-level evaluation of reasoning and actions, outcome-level evaluation
of the final result, and lifecycle-level evaluation of operational
properties. Consequently, an evaluation result can identify not only
whether an agent fails, but also the execution stage and capability
dimension associated with the failure.

\subsection{Extensible Evaluation}
\label{sec:extensible_metric_architecture}

The lifecycle taxonomy is designed as an organizational interface rather
than a closed metric checklist. Specifically, it separates
\emph{where a property is evaluated} from \emph{how that property is
measured}. Each metric is registered under an execution stage and a
fine-grained dimension, while its implementation may use an LLM judge, a
deterministic rule, an environment verifier, or a statistical aggregation
function. This separation allows heterogeneous evaluators to share a
consistent execution and reporting interface.

Such a design provides sustainable extensibility. When developers need to evaluate a new property, they can introduce a metric implementation and associate it with the corresponding lifecycle dimension without modifying
the benchmark runner, agent harness, or existing metrics. For example, a new
tool-selection metric can be added under the Tool dimension, while a new
latency or resource indicator can be added under Efficiency. Metrics may
also be benchmark-specific when a task exposes specialized execution
signals, while still remaining comparable through the shared taxonomy.

This extensible design follows the broader principle of treating evaluation
suites as continuously evolving systems rather than fixed collections of
scores. Similar modularity is adopted by holistic evaluation frameworks
that support the incremental addition of models, scenarios, and metrics
\citep{liang2022holistic}. In our framework, the taxonomy therefore provides
a stable high-level structure, whereas the metric catalog can evolve with
new agent capabilities, benchmarks, safety requirements, and deployment
conditions.

\subsection{Scalable Evaluation}
\label{sec:database_backed_evaluation}

Our evaluation is performed over task-level database records generated
during agent execution. Each run stores structured information describing
the task, agent configuration, interaction turns, model outputs, tool calls,
execution status, errors, resource usage, and other available trajectory
evidence. Metric evaluators subsequently query these records and write their
results back to the evaluation database.

Compared with storing each trajectory as an independent text or JSON file,
the database-backed design provides several engineering advantages. First,
a unified schema maintains explicit relationships among benchmarks, tasks,
runs, turns, tool calls, and metric results, reducing ambiguity introduced by
inconsistent file names or directory structures. Second, indexed queries
support efficient filtering and aggregation across models, harnesses,
benchmarks, run identifiers, and evaluation dimensions. Third, transactional
updates reduce the risk of partially written evaluation states and allow
failed or interrupted evaluations to be resumed more reliably.

More importantly, separating trajectory generation from metric computation
enables \emph{incremental evaluation}. Once an execution trajectory has been
stored, newly introduced metrics can be computed directly from the existing
database without rerunning the agent or repeating expensive API calls. The
same stored trajectory can therefore be evaluated under different metric
versions, judge models, or aggregation policies. Structured metadata also
improves experiment provenance, comparability, auditability, and
reproducibility, which are central requirements of production-oriented
machine-learning infrastructure
\citep{zaharia2018accelerating}.

The database primarily manages structured trajectory metadata and evaluation
results. Large auxiliary artifacts, when present, may remain in external
file or object storage, with their identifiers and locations recorded in the
database. This hybrid organization avoids using the database as an
inefficient large-file store while retaining centralized indexing,
traceability, and lifecycle management.


\section{Experiments}
\label{sec:experiments}
\newcommand{\bench}[1]{\texttt{#1}}
\newcommand{\benchref}[2]{\bench{#1}~\citep{#2}}

\newcommand{\harnesshead}[1]{%
  \makecell[c]{\scriptsize\sffamily\bfseries #1}%
}

\subsection{Harness--Benchmark Evaluation Matrix}
\label{subsec:harness-benchmark-matrix}

\begin{table*}[t]
\centering
\caption{Official benchmark performance across nine agent harnesses and two backbone models. Higher is better; \textbf{bold} marks the best harness in each row.}
\label{tab:harness-benchmark-matrix}
\small
\setlength{\tabcolsep}{3.2pt}
\renewcommand{\arraystretch}{1.12}
\begin{tabular*}{\textwidth}{@{}l@{\extracolsep{\fill}}*{9}{c}@{}}
\toprule
\textbf{Benchmark \& metric}
& \harnesshead{LangGraph}
& \harnesshead{CrewAI}
& \harnesshead{Google\\ADK}
& \harnesshead{AutoGen}
& \harnesshead{Smolagents}
& \harnesshead{Agno}
& \harnesshead{LlamaIndex}
& \harnesshead{Claude\\SDK}
& \harnesshead{OpenAI\\Agents} \\
\midrule
\rowcolor{Blue4Head!12}
\multicolumn{10}{@{}l@{}}{\rule{0pt}{2.2ex}\color{Blue4Head}\bfseries LLM: \texttt{gpt-5.6-sol}} \\
\makecell[l]{\textbf{DeepSearchQA}\\[-1pt]\scriptsize Paper F1 (\%)}
& 57.27 & 57.74 & \textbf{58.13} & 56.38 & 56.63 & 57.39 & 57.53 & 57.02 & 57.46 \\
\makecell[l]{\textbf{GDPval}\\[-1pt]\scriptsize Elo vs.\ human}
& \textbf{1129.6} & 1070.4 & 1053.9 & 1053.9 & 1085.1 & 1093.4 & 1063.3 & 1059.4 & 1041.4 \\
\makecell[l]{\textbf{$\tau$-bench}\\[-1pt]\scriptsize pass@1 (\%)}
& 60.78 & 35.29 & 76.79 & 77.94 & 77.91 & 76.39 & 55.88 & \textbf{84.72} & 80.56 \\
\makecell[l]{\textbf{$\tau^2$-bench}\\[-1pt]\scriptsize pass@1 (\%)}
& 63.73 & 37.86 & 73.47 & 73.53 & 81.11 & \textbf{82.67} & 65.69 & 81.08 & 81.94 \\
\makecell[l]{\textbf{$\tau^3$-bench}\\[-1pt]\scriptsize pass@1 (\%)}
& 61.39 & 38.61 & 79.25 & 76.47 & 82.22 & 80.00 & 62.50 & 83.78 & \textbf{84.00} \\
\makecell[l]{\textbf{Terminal-Bench 2.1}\\[-1pt]\scriptsize Success (\%)}
& 74.1 & 66.7 & 63.0 & 66.7 & \textbf{76.5} & 69.1 & 67.9 & 63.0 & 67.9 \\
\midrule
\rowcolor{Blue4Head!12}
\multicolumn{10}{@{}l@{}}{\rule{0pt}{2.2ex}\color{Blue4Head}\bfseries LLM: \texttt{glm-5.3}} \\
\makecell[l]{\textbf{DeepSearchQA}\\[-1pt]\scriptsize Paper F1 (\%)}
& 33.23 & 6.31 & 33.58 & 6.13 & \textbf{38.51} & 12.95 & 5.81 & 6.06 & 37.05 \\
\makecell[l]{\textbf{GDPval}\\[-1pt]\scriptsize Elo vs.\ human}
& 1061.0 & 1058.8 & 1032.0 & 1054.9 & 1024.5 & 1051.1 & \textbf{1070.4} & 1051.1 & 1054.9 \\
\makecell[l]{\textbf{$\tau$-bench}\\[-1pt]\scriptsize pass@1 (\%)}
& 52.17 & 15.65 & \textbf{63.48} & 42.61 & 62.61 & 45.22 & 42.61 & 58.26 & 45.22 \\
\makecell[l]{\textbf{$\tau^2$-bench}\\[-1pt]\scriptsize pass@1 (\%)}
& 57.89 & 23.68 & \textbf{62.28} & 46.49 & 58.77 & 43.86 & 42.98 & 59.65 & 43.86 \\
\makecell[l]{\textbf{$\tau^3$-bench}\\[-1pt]\scriptsize pass@1 (\%)}
& 47.37 & 20.18 & \textbf{63.16} & 45.61 & 50.88 & 42.98 & 46.49 & 57.89 & 43.86 \\
\makecell[l]{\textbf{Terminal-Bench 2.1}\\[-1pt]\scriptsize Success (\%)}
& 49.4 & 49.4 & 44.4 & 40.7 & 53.1 & 51.9 & \textbf{54.3} & 48.1 & 42.0 \\
\bottomrule
\end{tabular*}
\vspace{3pt}

\begin{minipage}{\textwidth}
\footnotesize
\textit{Notes.} DeepSearchQA reports paper F1; GDPval reports Bradley--Terry Elo against human gold (human $=1000$); the $\tau$ series reports Sierra \texttt{pass@1} on TRAJ-OK trajectories; and Terminal-Bench 2.1 reports resolved-task success. All harnesses use matched inference settings, tools, step limits, and timeout budgets within each backbone. Sample counts and filtering details appear in Appendix~\ref{app:exp-setup}.
\end{minipage}
\end{table*}

Table~\ref{tab:harness-benchmark-matrix} is not intended as a leaderboard of agent harnesses.
The purpose of this campaign is to show that \AtwoE{} can run, score, and compare many harnesses against many agent benchmarks and backbone models under a single pipeline.
The engine currently supports 23 benchmarks: \benchref{humaneval}{chen2021evaluating}, \benchref{swe-bench-lite}{jimenez2023swebench}, \benchref{swe-bench-verified}{openai2024swebenchverified}, \benchref{swe-bench-pro}{deng2025swe}, \benchref{agieval}{zhong2024agieval}, \benchref{arc-challenge}{clark2018think}, \benchref{bbh}{suzgun2023challenging}, \benchref{commonsenseqa}{talmor2019commonsenseqa}, \benchref{gsm8k}{cobbe2021training}, \benchref{hellaswag}{zellers2019hellaswag}, \benchref{math}{hendrycks2021math}, \benchref{mmlu}{hendrycks2020mmlu}, \benchref{mmlu-pro}{wang2024mmlupro}, \benchref{openbookqa}{mihaylov2018openbookqa}, \benchref{truthfulqa}{lin2021truthfulqa}, \benchref{traject-bench}{he2025trajectbench}, \benchref{DeepSearchQA}{gupta2026deepsearchqabridgingcomprehensivenessgap}, \benchref{gpqa}{rein2023gpqa}, \benchref{gdpval}{patwardhan2025gdpval}, \benchref{$\tau$-bench}{yao2024taubench}, \benchref{$\tau^2$-bench}{barres2025tau2bench}, \benchref{$\tau^3$-bench}{barres2026tau3bench}, and \benchref{terminal-bench-2.1}{merrill2026terminalbench}.
We report four of them---\benchref{DeepSearchQA}{gupta2026deepsearchqabridgingcomprehensivenessgap}, \benchref{gdpval}{patwardhan2025gdpval}, the $\tau$ series~\citep{yao2024taubench,barres2025tau2bench,barres2026tau3bench}, and \benchref{terminal-bench-2.1}{merrill2026terminalbench}---because they are the current frontier evaluations for deep web research, economically valuable professional work, conversational tool use, and hard command-line computer use.
The remaining registry entries are retained for coverage but are no longer the most discriminative agent tests.
The reported grid uses nine harnesses and two backbones (\texttt{gpt-5.6-sol} and \texttt{glm-5.3}).
Matched decoding, tools, step limits, timeouts, official scoring rules, and sample sizes are given in Appendix~\ref{app:exp-setup}.

The results reveal strong interactions among harness, benchmark, and model rather than a universal winner. Harness differences are modest on DeepSearchQA and GDPval with the stronger backbone, but become more pronounced with \texttt{glm-5.3}, while the $\tau$ series exhibits the largest variation across harnesses. Terminal-Bench 2.1 also shows a clear model effect, with \texttt{gpt-5.6-sol} achieving $68.3\%$ success compared with $48.1\%$ for \texttt{glm-5.3}. Overall, the leading harness changes across tasks and models, demonstrating that \AtwoE{} exposes task- and model-specific execution trade-offs rather than producing a single global ranking.
\begin{figure*}[t]
    \centering
    \includegraphics[
        width=\textwidth,
        keepaspectratio
    ]{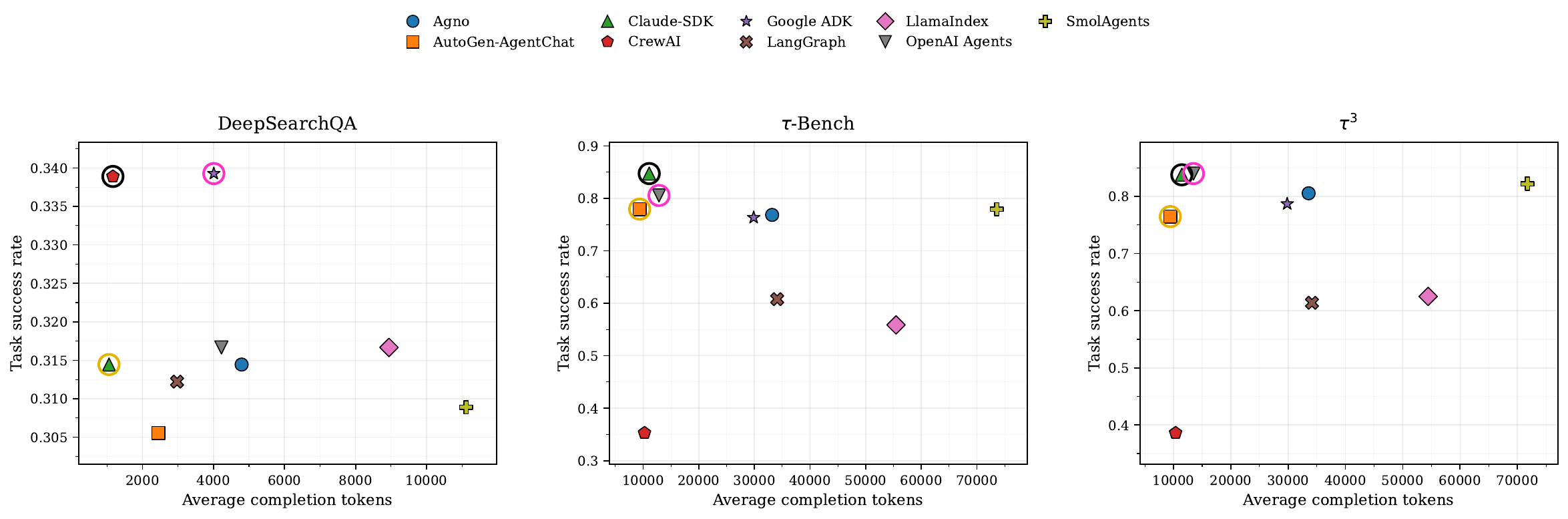}

    \caption{
    Comparison of nine agent harnesses across three benchmarks using
GPT-5.6-SOL as the common API model.
Each point represents one harness, with median total token usage on
the horizontal axis and task success rate on the vertical axis.
To jointly measure effectiveness and efficiency, we identify the top
three harnesses on each benchmark based on their trade-off between
achieving higher task success and using fewer tokens, and highlight
them using rank-specific colored circles, where black, magenta, and
gold circles denote the first-, second-, and third-ranked harnesses,
respectively.
The top-ranked harnesses are CrewAI, Google ADK, and Claude-SDK
on DeepSearchQA; Claude-SDK, OpenAI Agents, and AutoGen-AgentChat
on $\tau$-Bench; and Claude-SDK, OpenAI Agents, and
AutoGen-AgentChat on $\tau^3$-bench.
These results demonstrate that our evaluation framework enables unified
cross-benchmark and cross-harness analysis, revealing substantial
differences in both task effectiveness and token efficiency under a
consistent model backend and evaluation protocol.
    }

    \label{fig:cross_benchmark_harness_comparison}
\end{figure*} 
\subsection{Cross-Benchmark Harness Comparison}
\label{subsec:Cross-Benchmark-Harness-Comparison}
Figure~\ref{fig:cross_benchmark_harness_comparison} presents the performance of nine agent harnesses across three representative benchmarks: DeepSearchQA, $\tau$-Bench, and $\tau^3$-bench. All experiments use GPT-5.6-SOL as the underlying API model, providing a controlled and consistent model backend for comparing different harness implementations. In each subplot, the horizontal axis represents token usage, while the vertical axis reports the task success rate. The visualization therefore jointly reflects task effectiveness and execution efficiency, rather than evaluating agent systems solely based on final-task accuracy.

To jointly compare effectiveness and efficiency, we score each harness according to its distance from the desirable region of high task success and low token usage. For each benchmark, token usage and task success rate are normalized across all harnesses. The score is defined as:

\begin{equation}
Q_h =
1-
\frac{
\sqrt{
\hat{T}_h^2+
(1-\hat{S}_h)^2
}
}{\sqrt{2}},
\end{equation}

where $\hat{T}_h$ and $\hat{S}_h$ denote the normalized token usage and task success rate of harness $h$, respectively. A higher $Q_h$ indicates a better balance between effectiveness and efficiency. Based on this score, the top three harnesses on each benchmark are highlighted using rank-specific colored circles, where black, magenta, and gold denote the first-, second-, and third-ranked harnesses, respectively.

Despite using the same underlying model, the evaluated harnesses occupy substantially different positions in the effectiveness--efficiency space. These differences are reflected not only in task success rates but also in the amount of token usage required to complete the same benchmark. This demonstrates that an agent harness is not merely a lightweight wrapper around an API model. Differences in prompt construction, tool representation, context management, execution loops, error handling, and termination policies can lead to substantially different execution behaviors even when the underlying model is fixed.

The figure also shows that the relative performance of harnesses depends strongly on the benchmark. On DeepSearchQA, success rates are relatively concentrated across harnesses, while token usage varies by nearly an order of magnitude. Consequently, efficiency differences become particularly important for distinguishing harnesses on this benchmark. CrewAI achieves the highest effectiveness--efficiency score, followed by Google ADK and Claude-SDK.

In contrast, both $\tau$-Bench and $\tau^3$-bench exhibit considerably larger variation in task success rate as well as token consumption. On $\tau$-Bench, Claude-SDK, OpenAI Agents, and AutoGen-AgentChat achieve the three highest effectiveness--efficiency scores. The same three harnesses also rank highest on $\tau^3$-bench, with Claude-SDK ranked first, OpenAI Agents second, and AutoGen-AgentChat third. Notably, several harnesses consume substantially more tokens without achieving correspondingly higher task success, illustrating that increased execution cost does not necessarily translate into better task performance.

Overall, the clear separation among harnesses demonstrates the discriminative capability of our agent evaluation framework. Even under an identical GPT-5.6-SOL model backend, the framework captures meaningful differences in effectiveness and execution efficiency introduced by different harness implementations. Rather than producing nearly identical evaluations for systems sharing the same model, our evaluation reveals how harness-level design choices influence the extent to which underlying model capabilities are translated into practical agent performance.
\begin{table*}[t]
    \centering
    \small
    \setlength{\tabcolsep}{5pt}
    \renewcommand{\arraystretch}{1.12}

    \begin{tabularx}{\textwidth}{
        >{\raggedright\arraybackslash}p{0.15\textwidth}
        >{\raggedright\arraybackslash}X
        >{\raggedright\arraybackslash}X}

        \toprule
        \textbf{Dimension} &
        \textbf{Google-ADK (early stall)} &
        \textbf{LangGraph (verification gap)} \\
        \midrule

        Task outcome
        &
        Stops after repository/environment inspection.
        NumPy patching, build, installation, README verification, and tests
        are never attempted.
        \newline
        \textbf{Metrics:}
        \texttt{task\_completion}=0,
        \texttt{correctness}=0
        &
        Patches NumPy compatibility, builds and installs the extensions,
        and verifies the README example, but times out before
        \texttt{pytest tests/}.
        \newline
        \textbf{Metrics:}
        \texttt{task\_completion}=0,
        \texttt{correctness}=0
        \\

        \midrule

        Execution behavior
        &
        6 \texttt{bash} calls, all used for reconnaissance; the trajectory
        never enters the build/install loop.
        \newline
        \textbf{Metrics:}
        \texttt{tool\_call\_count}=6,
        \texttt{repeated\_tool\_call\_rate}=0
        &
        22 tool steps covering patching, rebuilding, installation, and
        README verification; execution stops during final test preparation.
        \newline
        \textbf{Metrics:}
        \texttt{tool\_call\_count}=22,
        \texttt{repeated\_tool\_call\_rate}=0
        \\

        \midrule

        Tool interaction
        &
        All tool calls are schema-valid, but none perform the required
        engineering operations.
        \newline
        \textbf{Metric:}
        \texttt{tool\_invocation}=1.0
        &
        Most calls are valid, but the final \texttt{bash} invocation is
        truncated/malformed.
        \newline
        \textbf{Metric:}
        \texttt{tool\_invocation}=0.0
        \\

        \midrule

        Planning diagnosis
        &
        Goal direction is correct, but the plan is shallow and incomplete:
        inspection never transitions into execution.
        \newline
        \textbf{Metrics:}
        \texttt{plan\_goal\_alignment}=1.0,
        \texttt{plan\_completeness}=0.0,
        \texttt{plan\_grade}=0.2
        &
        The plan is technically strong and most required engineering steps
        are completed, but the final verification phase is missing.
        \newline
        \textbf{Metrics:}
        \texttt{plan\_goal\_alignment}=1.0,
        \texttt{plan\_completeness}=0.0,
        \texttt{plan\_grade}=1.0
        \\

        \midrule

        Failure reporting
        &
        Times out without reporting incomplete progress or the missing
        execution phases.
        \newline
        \textbf{Metric:}
        \texttt{failure\_transparency}=0.0
        &
        Times out after partial success without reporting that the README
        check passed or that repository tests remain unverified.
        \newline
        \textbf{Metric:}
        \texttt{failure\_transparency}=0.0
        \\

        \midrule

        Overall diagnosis
        &
        \textbf{Early plan-depth failure:}
        correct reconnaissance, but no transition to execution.
        &
        \textbf{Late verification failure:}
        near-complete engineering trajectory, but tests are not executed
        and the final tool call is malformed.
        \\

        \bottomrule

    \end{tabularx}

    \caption{
    Metric-based comparison of two GPT-5.6-sol Terminal Bench~2.1
    trajectories on the same \texttt{pyknotid} task.
    Although both runs receive zero task completion and correctness,
    the diagnostic metrics distinguish an early plan-depth failure in
    Google-ADK from a late verification and tool-invocation failure in
    LangGraph.
    }
    \label{tab:tb21_pyknotid_metric_case_study}
\end{table*}

\subsection{Case Study}
\label{subsec:case-study}
Despite using the same API model, different harnesses show large gaps in both
success rate and efficiency. On our campaign runs, the harness-level success-rate
spread is substantial across benchmarks, while token use and wall time also move
considerably between implementations. These patterns suggest that the harness is
not a thin wrapper around the model: it shapes prompt construction, tool routing,
state handling, and when the run stops. More importantly, the evaluator can tell
harnesses apart even when the backend model is fixed---instead of collapsing
every system on the same API into nearly identical scores.
To isolate harness effects, we compare two complete trajectories from the
\textbf{same API model} (GPT-5.6-sol) on the \textbf{same Terminal Bench~2.1
task}---Task~41, compiling \texttt{pyknotid} with NumPy~2.3.0 compatibility.
Both trajectories fail the benchmark verifier
(\texttt{correctness}=0, \texttt{task\_completion}=0), so a
correctness-only view would treat them as the same outcome. Our engine does
not stop there.
Google-ADK (run~44) makes six \texttt{bash} calls---clone, read
\texttt{setup.py} and \texttt{.pyx} files, scan for deprecated NumPy aliases,
check Cython/gcc---and then stops. No patch, no \texttt{build\_ext}, no
\texttt{pip install}, no README check, no \texttt{pytest}. The harness times
out with an empty \texttt{final\_answer}.
LangGraph (run~30) goes much further on the same prompt: it installs Cython,
patches NumPy~2.x aliases, builds extensions, fixes a Python~3.13 \texttt{gcd}
import, and passes the README snippet (Alexander polynomial $\approx 7.0$).
It still fails because it never runs the required \texttt{tests/} suite and
dies mid-reinstall without telling the user what finished and what did not.
The metric breakdown makes this gap explicit rather than hiding it behind a
shared zero.
On planning, Google-ADK gets \texttt{plan\_grade}=0.2
(\emph{mostly\_incorrect}) while LangGraph gets \texttt{plan\_grade}=1.0
(\emph{perfect}); both have \texttt{plan\_completeness}=0.0, but for
different reasons---early abandonment versus stopping one verification step
short.
On tools, the picture flips: Google-ADK keeps \texttt{tool\_invocation}=1.0
because every call it made was a valid simple \texttt{bash} invocation;
LangGraph drops to \texttt{tool\_invocation}=0.0 after the judge flags a
truncated shell command in the final segment, even though most of the
engineering chain was sound.
Both score \texttt{failure\_transparency}=0.0 (\emph{opaque}): neither
explains at timeout what was done, what remains, or---in LangGraph's
case---that the README demo had already succeeded.
Because model, task, and environment are fixed, these differences reflect
harness-level execution behavior, not model capability noise.
Google-ADK treats the task as an open-ended survey and never enters the
build/install loop.
LangGraph does real work but spends budget on polish and exits before the
test gate, without a clear final status.
That is the point of discriminative evaluation on a fixed API backend: the
engine separates shallow planning from near-complete execution, clean simple
tool use from a broken final invocation, and shows \emph{why} two failed runs
should not be grouped together.

\newpage

\bibliographystyle{colm2025_conference}
\bibliography{custom}

\newpage
\appendix
\section{Experimental Setup}
\label{app:exp-setup}

The campaign in \S\ref{subsec:harness-benchmark-matrix} uses nine registered harnesses---LangGraph, CrewAI, Google~ADK, AutoGen AgentChat, Smolagents, Agno, LlamaIndex, Claude SDK, and OpenAI Agents---and two backbone models, \texttt{gpt-5.6-sol} and \texttt{glm-5.3}.
Within each backbone, all harnesses share the same task IDs, inference configuration, tool setup, step limit, and timeout budget.
Each completed task is scored with its benchmark's official protocol.

The engine defaults are eight turns, 4{,}096 output tokens per completion, and a 180-second timeout for each LLM call.
Official full-suite scripts override these values by benchmark, and the registry-level agent overrides set the turn limit to 30 for the $\tau$ series and to eight for DeepSearchQA and GDPval-AA.
Terminal-Bench 2.1 instead inherits its execution limits directly from each benchmark task.
Table~\ref{tab:official-run-budgets} gives the resulting per-run budgets.
For the three A\textsuperscript{2}E-native benchmark families, every LLM call is retried at most twice.
Neither \texttt{temperature} nor \texttt{top\_p} is set explicitly for these runs; requests therefore use the gateway and provider-SDK defaults.

\begin{table}[H]
\centering
\caption{Per-run budgets used in the evaluation. Max tokens denotes the limit for one model completion.}
\label{tab:official-run-budgets}
\small
\setlength{\tabcolsep}{4pt}
\renewcommand{\arraystretch}{1.15}
\begin{tabular*}{\textwidth}{@{}l@{\extracolsep{\fill}}cccccl@{}}
\toprule
\textbf{Benchmark family}
& \makecell{\textbf{Turns /}\\\textbf{steps}}
& \makecell{\textbf{Max}\\\textbf{tokens}}
& \makecell{\textbf{LLM}\\\textbf{timeout}}
& \makecell{\textbf{Harness}\\\textbf{timeout}}
& \makecell{\textbf{Run}\\\textbf{deadline}}
& \textbf{Permitted tools or context} \\
\midrule
$\tau$ / $\tau^2$ / $\tau^3$ & 30 & 4{,}096 & 180\,s & 1{,}200\,s & 1{,}100\,s & Sierra retail tools and task DB \\
DeepSearchQA & 8 & 4{,}096 & 180\,s & 720\,s & 620\,s & \texttt{web\_search}, \texttt{open\_url} \\
GDPval-AA & 8 & 16{,}384 & 600\,s & 1{,}800\,s & 1{,}700\,s & No tools; file-backed task context \\
Terminal-Bench 2.1 & n/a & n/a & n/a & Per task & n/a & Shell in the official task container \\
\bottomrule
\end{tabular*}
\vspace{2pt}

\begin{minipage}{\textwidth}
\footnotesize
Tool returns are truncated to 2{,}500 characters by default; $\tau$ leftover recovery runs use an 8{,}000-character limit. Terminal-Bench 2.1 does not impose a suite-wide turn, token, or per-call LLM limit.
\end{minipage}
\end{table}

Terminal-Bench 2.1~\citep{terminalbench2026v21} defines its budgets independently for every task in the official \texttt{task.toml} files.
We apply no global timeout or resource override: each trial uses the specified agent timeout, verifier timeout, container build timeout, CPU count, memory, storage, GPU count, and network policy.
Across the released 89-task suite, agent timeouts range from 600 to 12{,}000 seconds and verifier timeouts from 360 to 12{,}000 seconds.
All environments use a 600-second build timeout and 10{,}240\,MB of storage; task allocations range from one to four CPUs and from 2{,}048 to 8{,}192\,MB of memory.
The official configurations allocate no GPUs and permit Internet access.
We exclude eight network-security tasks because \texttt{gpt-5.6-sol} refuses to answer them; to preserve matched task IDs across backbones, we remove the same tasks from both model evaluations, yielding a fixed set of 81 tasks.
For all included tasks, we preserve the corresponding task-level values exactly and run each task in its pinned benchmark container.

\end{document}